\documentclass[11pt,a4paper]{article}
\pdfoutput=1
\usepackage[hyphens]{url}
\usepackage[hyperref]{acl2018}
\usepackage{times}
\usepackage{latexsym}

\aclfinalcopy 

\usepackage{amsmath,amssymb}
\usepackage{multirow}
\usepackage{color}
\usepackage{graphicx}
\usepackage{bbm}
\usepackage{xspace}
\usepackage{wasysym}
\usepackage{algorithmic}
\usepackage{float}
\usepackage{mathtools}
\usepackage{pgfplots}
\pgfplotsset{compat=1.14}
\usepackage{placeins}
\usepackage{tabularx}
\usepackage{booktabs}
\usepgfplotslibrary{groupplots}
\usepackage{xcolor}
\usepackage{colortbl}
\aclfinalcopy

\newcolumntype{L}[1]{>{\raggedright\arraybackslash}p{#1}}
\newcolumntype{C}[1]{>{\centering\arraybackslash}p{#1}}
\newcolumntype{R}[1]{>{\raggedleft\arraybackslash}p{#1}}\mathtoolsset{showonlyrefs}

\newcommand{\sent}{s}
\newcommand{\sentt}{s_t}

\newcommand{\paranmtnonum}{\textsc{ParaNMT}\xspace}
\newcommand{\paranmt}{\textsc{ParaNMT-50M}\xspace}

\newcommand{\paragramsl}{\textsc{paragram-sl999}\xspace}
\newcommand{\paragramphrase}{\textsc{paragram-phrase}\xspace}

\newcommand{\lstmavg}{\textsc{LSTM}\xspace}
\newcommand{\blstmavg}{\textsc{BLSTM}\xspace}
\newcommand{\wordavg}{\textsc{Word}\xspace}
\newcommand{\triavg}{\textsc{Trigram}\xspace}
\newcommand{\concat}{,\xspace}

\newcommand{\stddev}[1]{\tiny{$\pm$#1}}

\DeclareMathOperator*{\argmax}{argmax}

\title{\paranmt: Pushing the Limits of Paraphrastic Sentence Embeddings with Millions of Machine Translations}

\author{John Wieting$^1$\ \ \ \ \ \ Kevin Gimpel$^2$\\
$^1$Carnegie Mellon University,
Pittsburgh, PA, 15213, USA\\
$^2$Toyota Technological Institute at Chicago, Chicago, IL, 60637, USA\\
\tt{jwieting@cs.cmu.edu},  \tt{kgimpel@ttic.edu}
}

\date{}

\begin{document}
\maketitle
\begin{abstract}
We describe \paranmt, a dataset of more than 50 million English-English sentential paraphrase pairs. We generated the pairs automatically by using neural machine translation to translate the non-English side of a large parallel corpus, following \citet{wieting2017backtrans}. 
Our hope is that \paranmt can be a valuable resource for paraphrase generation and can provide a rich source of semantic knowledge to improve downstream natural language understanding tasks. 
To show its utility, we use \paranmt to train paraphrastic sentence embeddings that outperform all supervised systems on every SemEval semantic textual similarity competition, in addition to 
showing how it can be 
used  for paraphrase generation.\footnote{
Dataset, code, and embeddings are available at 
\url{https://www.cs.cmu.edu/~jwieting}.} 

\end{abstract}

\section{Introduction}
While many approaches have been developed for generating or finding paraphrases~\cite{barzilay2001extracting,dolan2004unsupervised,lan-EtAl:2017:EMNLP20171}, there do not exist any freely-available datasets with millions of sentential paraphrase pairs. 
The closest such resource is the Paraphrase Database (PPDB; \citealp{GanitkevitchDC13}), which was created automatically from bilingual text by pivoting over the non-English language~\citep{bannard2005paraphrasing}. PPDB has been used to improve word embeddings~\citep{faruqui-15,mrkvsic2016counter}. However, PPDB is less useful for learning \emph{sentence} embeddings~\cite{wieting-17-full}. 

In this paper, we describe the creation of a dataset containing more than 50 million sentential paraphrase pairs.  
We create it automatically by scaling up the approach of \citet{wieting2017backtrans}. We use neural machine translation (NMT) to translate the Czech side of a large Czech-English parallel corpus. We pair the English translations with the English references to form paraphrase pairs. 
We call this dataset \paranmt. 
It contains examples illustrating a broad range of paraphrase phenomena; we show examples in Section~\ref{sec:data}. 
\paranmt has the potential 
to be useful for many tasks, from linguistically controlled paraphrase generation, style transfer, and sentence simplification to core NLP problems like machine translation. 

We show the utility of \paranmt by using it to train  
paraphrastic sentence embeddings using the learning framework of \citet{wieting-16-full}. 
We primarily evaluate our sentence embeddings on the SemEval semantic textual similarity (STS) competitions 
from 2012-2016. 
Since so many domains are covered in these datasets, they form a demanding evaluation for a general purpose sentence embedding model.

Our sentence embeddings learned from \paranmt outperform all systems in every STS competition from 2012 to 2016. 
These tasks have drawn substantial participation; 
in 2016, for example, the competition attracted 43 teams and had 119 submissions. 
Most STS systems use curated lexical resources, the provided supervised training data with manually-annotated similarities, and joint modeling of the sentence pair. We use none of these, simply encoding each sentence independently using our models and computing cosine similarity between their embeddings.  

We experiment with several compositional architectures and find them all to work well. We also find benefit 
from making a simple change to learning to better leverage the large training set, namely, increasing the search space of negative examples. We additionally evaluate on general-purpose sentence embedding tasks used in past work~\citep{kiros2015skip,conneau2017supervised}, finding our embeddings to perform competitively.

Lastly, we show that \paranmt is able to be used in paraphrase generation. Recent work~\cite{iyyer2018controlled} used \paranmt to generate paraphrases that have a specific syntactic form. In their model, a sentence and its target form (represented as the top two levels of a linearized parse tree) are transformed by the model into a paraphrase with this target structure. We also explore paraphrase generation in this paper, finding that a basic encoder-decoder model trained on \paranmt has a canonicalization effect and is able to correct grammar and standardize the input sentence.

We release the \paranmt dataset, our trained sentence embeddings, and our code. 
\paranmt is the largest collection of sentential paraphrases released to date. We hope it can motivate new research directions and be used to create powerful NLP models, while adding a robustness to existing ones by incorporating paraphrase knowledge. 
Our paraphrastic sentence embeddings are state-of-the-art by a significant margin, and we hope they can be useful for many applications both as a sentence representation function and as a general similarity metric.

\section{Related Work}
We discuss work in automatically building paraphrase corpora, learning general-purpose sentence embeddings, and using parallel text for learning embeddings and similarity functions. 

\paragraph{Paraphrase discovery and generation.} 
Many methods have been developed for generating or finding paraphrases, including using multiple  translations of the same source material~\cite{barzilay2001extracting}, using comparable articles from multiple news sources~\cite{dolan2004unsupervised,dolan-05,quirk-04}, aligning sentences between standard and Simple English Wikipedia~\cite{coster2011simple}, crowdsourcing~\cite{Xu:ea:2014,xu2015semeval,jiang-kummerfeld-lasecki:2017:Short}, using diverse MT systems to translate a single source sentence~\cite{suzuki-kajiwara-komachi:2017:ACL-2017-Student-Research-Workshop}, and using tweets with matching URLs~\cite{lan-EtAl:2017:EMNLP20171}. 

The most relevant prior work uses bilingual corpora. \newcite{bannard2005paraphrasing} used methods from statistical machine translation to find lexical and phrasal paraphrases in parallel text. \citet{GanitkevitchDC13} scaled up these techniques to produce the Paraphrase Database (PPDB). 
Our goals are similar to those of PPDB, which has likewise been generated for many languages~\cite{ganitkevitch2014multilingual} since it only needs parallel text. 
In particular, we follow the approach of \citet{wieting2017backtrans}, who used NMT to translate the non-English side of parallel text to
get English-English paraphrase pairs. 
We 
scale up the method to a larger dataset, produce state-of-the-art paraphrastic sentence embeddings, and release all of our resources.

\paragraph{Sentence embeddings.} 
Our learning and evaluation setting is the same as that in recent work which seeks to learn paraphrastic sentence embeddings that can be used for downstream tasks~\cite{wieting-16-full, wieting2016charagram, wieting-17-full, wieting2017backtrans}. They trained models on noisy paraphrase pairs and evaluated them primarily on semantic textual similarity (STS) tasks. 
Prior work in learning general sentence embeddings has used  autoencoders~\cite{SocherEtAl2011:PoolRAE,hill2016learning}, 
encoder-decoder architectures~\cite{kiros2015skip,gan-EtAl:2017:EMNLP2017}, and other learning frameworks~\cite{le2014distributed,pham-EtAl:2015:ACL-IJCNLP,arora2017simple,pagliardini2017unsupervised,conneau2017supervised}.

\paragraph{Parallel text for learning embeddings.} 

Prior work has shown that parallel text, and resources built from parallel text like NMT systems and PPDB, can be used for learning embeddings for words and sentences. 
Several have used PPDB as a knowledge resource for training or improving embeddings~\cite{faruqui-15,wieting2015paraphrase,mrkvsic2016counter}. 
Several have used NMT architectures and training settings to obtain better embeddings for words~\cite{hill2014embedding,hill-14} and words-in-context~\cite{mccann2017learned}. 
\newcite{hill2016learning} evaluated the encoders of English-to-X NMT systems as sentence representations. 
\newcite{E17-1083} adapted trained NMT models to produce sentence similarity scores in semantic evaluations.

\section{The \paranmt Dataset} \label{sec:data}
 
\begin{table*}[t]
\setlength{\tabcolsep}{4pt}
\centering
\small
\begin{tabular} { | l || c | c | c | c | c | c |}
\hline
Dataset & Avg.~Length & Avg.~IDF & Avg.~Para.~Score & Vocab.~Entropy & Parse Entropy & Size \\
\hline
Common Crawl & 24.0\stddev{34.7} & 7.7\stddev{1.1} & 0.83\stddev{0.16} & 7.2 & 3.5 & 0.16M \\
CzEng 1.6 & 13.3\stddev{19.3} & 7.4\stddev{1.2} & 0.84\stddev{0.16} & 6.8 & 4.1 & 51.4M \\
Europarl & 26.1\stddev{15.4} & 7.1\stddev{0.6} & 0.95\stddev{0.05} & 6.4 & 3.0 & 0.65M \\
News Commentary & 25.2\stddev{13.9} & 7.5\stddev{1.1} & 0.92\stddev{0.12} & 7.0 & 3.4 & 0.19M \\
\hline
\end{tabular}
\caption{\label{table:CorpusStsts}
Statistics of 100K-samples of Czech-English parallel corpora;  
standard deviations are shown for averages.  
}
\end{table*}

\begin{table*}[t]
\setlength{\tabcolsep}{4pt}
\small
\centering
\begin{tabular} { | l | l | } 
\hline
Reference Translation & Machine Translation \\
\hline
so, what's half an hour? & half an hour won't kill you. \\
well, don't worry. i've taken out tons and tons of guys. lots of guys. &	don't worry, i've done it to dozens of men. \\
it's gonna be ...... classic. & yeah, sure. it's gonna be great. \\
greetings, all! &	hello everyone! \\
but she doesn't have much of a case. & but as far as the case goes, she doesn't have much. \\
it was good in spite of the taste. &	despite the flavor, it felt good. \\
\hline
\end{tabular}
\caption{\label{table:paranmt-examples}
Example paraphrase pairs from \paranmt, where each consists of an English reference translation and the machine translation of the Czech source sentence (not shown).
}
\end{table*}

To create our dataset, we used back-translation~\citep{wieting2017backtrans}. 
We used a Czech-English NMT system to translate Czech sentences from the training data into English. We paired the translations with the English references to form English-English paraphrase pairs. 

We used the pretrained Czech-English model from the NMT system of \citet{sennrich2017nematus}. Its training data includes four sources: Common Crawl, CzEng 1.6~\cite{czeng16:2016}, Europarl, and News Commentary. 
We next discuss how we chose the CzEng corpus from among these to create our dataset. 
We did not choose Czech due to any particular linguistic properties. \citet{wieting2017backtrans} found little difference among Czech, German, and French as source languages for back-translation. There were much larger differences due to data domain, so we focus on the question of domain in this section. 
We leave the question of investigating properties of back-translation of different languages to future work. 

\subsection{Choosing a Data Source}
To assess characteristics that yield useful data, we randomly sampled 100K English reference translations from each data source and computed statistics. Table~\ref{table:CorpusStsts} shows the average sentence length, the average inverse document frequency (IDF) where IDFs are computed using Wikipedia sentences, and the average paraphrase score for the two sentences. 
The paraphrase score 
is calculated by averaging \paragramphrase embeddings~\cite{wieting-16-full} for the two sentences in each pair and then computing their cosine similarity. 
The table also shows the entropies of the vocabularies and constituent parses obtained using the Stanford Parser~\cite{manning-EtAl:2014:P14-5}.\footnote{To mitigate sparsity in the parse  entropy, we used only the top two levels of each parse tree.} 

Europarl exhibits the least diversity in terms of rare word usage, vocabulary entropy, and parse entropy. This is unsurprising given its formulaic and repetitive nature. 
CzEng has shorter sentences than the other corpora and more diverse sentence structures, as shown by its high parse entropy. 
In terms of vocabulary use, CzEng is not particularly more diverse than Common Crawl and News Commentary, though this could be due to the prevalence of named entities in the latter two. 

In Section~\ref{sec:data-comp}, we empirically compare these data sources as training data for sentence embeddings. The CzEng corpus yields the strongest performance when controlling for training data size. 
Since its sentences are short, we suspect this helps ensure high-quality back-translations. 
A large portion of it is movie subtitles which tend to use a wide vocabulary and have a diversity of sentence structures; however, other domains are included as well. 
It is also the largest corpus, containing over 51 million sentence pairs. In addition to providing a large number of training examples for downstream tasks, this means that the NMT system should be able to produce quality translations for this subset of its training data. 

For all of these reasons, we chose the CzEng corpus to create \paranmt. When doing so, we used beam search with a beam size of 12 and selected the highest scoring translation from the beam. It took over 10,000 GPU hours to back-translate the CzEng corpus. 
We show illustrative examples in Table~\ref{table:paranmt-examples}. 

\subsection{Manual Evaluation} \label{sec:human}

We conducted a manual analysis of our dataset in order to quantify its noise level, and how the noise can be ameliorated with filtering. 
Two domain experts annotated a sample of 100 examples from each of five ranges of the Paraphrase Score.\footnote{Since the range of values is constrained to be $\leq 1$, and most values are positive, we split it up into 5 evenly spaced segments as shown in Table~\ref{table:human}.} 
They annotated both the strength of the paraphrase relationship and the fluency of the back-translation.

\begin{table}[h!]
\setlength{\tabcolsep}{4pt}
\small
\centering
\begin{tabular} { | c | r | c | c  c  c | c  c  c |} 
\hline
\multicolumn{1}{|c|}{Para.~Score} & \multicolumn{1}{c|}{\#} & \multicolumn{1}{c|}{Tri. Overlap} & \multicolumn{3}{c|}{Paraphrase} & \multicolumn{3}{c|}{Fluency} \\
Range & \multicolumn{1}{c|}{(M)} & Mean (Std.) & 1 & 2 & 3 & 1 & 2 & 3  \\
\hline
(-0.1, 0.2] 
& 4.0 & 0.00\stddev{0.0} 
& 92 & 6 & 2 & 1 & 5 & 94 \\
(0.2, 0.4] 
& 3.8 & 0.02\stddev{0.1} 
& 53 & 32 & 15 & 1 & 12 & 87 \\
(0.4, 0.6] 
& 6.9 & 0.07\stddev{0.1} 
& 22 & 45 & 33 & 2 & 9 & 89 \\
(0.6, 0.8] 
& 14.4 & 0.17\stddev{0.2} 
& 1 & 43 & 56 & 11 & 0 & 89 \\
(0.8, 1.0] 
& 18.0 & 0.35\stddev{0.2} 
& 1 & 13 & 86 & 3 & 0 & 97 \\
\hline
\end{tabular}
\caption{\label{table:human}
Manual evaluation of 100-pair data samples drawn from five ranges of the automatic paraphrase score (first column). Second column shows total count of pairs in that range in \paranmt. Paraphrase strength and fluency were judged on a 1-3 scale and the table shows counts of each score designation. 
}
\end{table}

To annotate paraphrase strength, we adopted the annotation guidelines used by \citet{agirre2012semeval}. The original guidelines specify 6 classes, which we reduce to 3 for simplicity. We collapse the top two into one category, leave the next alone, and collapse the bottom 3 into our lowest category. Therefore, for a sentence pair to have a rating of 3, the sentences must have the same meaning, but some unimportant details can differ. To have a rating of 2, the sentences are roughly equivalent, with some important information missing or that differs slightly. For a rating of 1, the sentences are not equivalent, even if they share minor details.

For fluency of the back-translation, we use the following: A rating of 3 means it has no grammatical errors, 2 means it has one to two errors, and 1 means it has more than two grammatical errors or is not a natural English sentence.

Table~\ref{table:human} summarizes the annotations. For each score range, we report the number of pairs, the mean trigram overlap score, and the number of times each paraphrase/fluency label was present in the sample of 100 pairs. 
There is noise in the dataset but it is largely confined in the bottom two ranges which together comprise only 16\% of the entire dataset. In the highest paraphrase score range, 86\% of the pairs possess a strong paraphrase relationship. The annotations suggest that \paranmt contains approximately 30 million strong paraphrase pairs, and that the paraphrase score is a good indicator of quality. 
With regards to fluency, the vast majority of the back-translations are fluent, even at the low end of the paraphrase score range. At the low ranges, we inspected the data and found there to be many errors in the sentence alignment in the original bitext.

\section{Learning Sentence Embeddings} \label{sec:models}
To show the usefulness of the \paranmt dataset, we will use it to train sentence embeddings. 
We adopt the learning framework from \citet{wieting-16-full}, which was developed to train sentence embeddings from pairs in PPDB. We first describe the compositional sentence embedding models we will experiment with, then discuss training and our modification (``mega-batching''). 

\paragraph{Models.} 
We want to embed a word sequence $\sent$ into a fixed-length vector. 
We denote the $t$th word in $\sent$ as $\sentt$, and we denote its word embedding by $x_t$. 
We focus on three model families, though we also experiment with combining them in various ways.
The first, which we call \wordavg, simply averages the embeddings $x_t$ of all words in $\sent$. 
This model was found by \newcite{wieting-16-full} to perform strongly for semantic similarity tasks. 

The second is similar to \wordavg, but instead of word embeddings, we average character trigram embeddings~\citep{huang2013learningshort}. We call this \triavg. 
\citet{wieting2016charagram} found this to work well for sentence embeddings compared to other $n$-gram orders and to word averaging. 

The third family includes long short-term memory (LSTM) architectures~\citep{hochreiter1997long}. We 
average the hidden states to produce the final sentence embedding. 
For regularization during training, we scramble words with a small probability~\citep{wieting-17-full}. 
We also experiment with bidirectional LSTMs (\blstmavg), averaging the forward and backward hidden states with no concatenation.\footnote{Unlike \citet{conneau2017supervised}, we found this to outperform max-pooling for both semantic similarity and general sentence embedding tasks.} 

\paragraph{Training.} 
The training data is a set $S$ 
of paraphrase pairs $\langle s, s'\rangle$ and 
we minimize a margin-based loss $\ell(s, s') =$
\begin{align}
\max(0,\delta - \cos(g(s), g(s')) + \cos(g(s), g(t)))
\label{eq:obj}
\end{align} 
\noindent where $g$ is the model (\wordavg, \triavg, etc.), 
$\delta$ is the margin, 
and 
$t$ is a ``negative example'' taken from a mini-batch during optimization. 
The intuition is that we want the two texts to be more similar to each other than to their negative examples. 
To select $t$
we choose the most similar sentence in some set.
For simplicity 
we use the mini-batch for this set, i.e.,
\begin{equation}
t = \argmax_{t' : \langle t', \cdot\rangle \in S_b \setminus \{\langle s, s'\rangle\}} \cos(g(s), g(t'))
\end{equation}
\noindent where $S_b\subseteq S$ is the current mini-batch. 

\paragraph{Modification: mega-batching.} 
By using the mini-batch to select negative examples, we may be limiting the learning procedure. That is, if all potential negative examples in the mini-batch are highly dissimilar from $s$, the loss will be too easy to minimize. 
Stronger negative examples can be obtained by using larger mini-batches, but large mini-batches are sub-optimal for optimization. 

Therefore, we propose a procedure we call ``mega-batching.'' We aggregate $M$ mini-batches to create one mega-batch and select negative examples from the mega-batch. 
Once each pair in the mega-batch has a negative example, the mega-batch is split back up into $M$ mini-batches and training proceeds. We found that this provides more challenging negative examples during learning as shown in Section~\ref{sec:megabatch}. Table~\ref{table:STSresultsMB} shows results for different values of $M$, showing consistently higher correlations with larger $M$ values.

\section{Experiments}  \label{sec:exp}
We now investigate how best to use our generated paraphrase data for training paraphrastic sentence embeddings. 

\subsection{Evaluation}

We evaluate sentence embeddings using the 
SemEval semantic textual similarity (STS) tasks from 2012 to 2016~\cite{agirre2012semeval,diab2013eneko,agirre2014semeval,agirre2015semeval,agirre2016semeval} and the STS Benchmark~\cite{cer2017semeval}.  
Given two sentences, the aim of the STS tasks is to predict their similarity on a 0-5 scale, where 0 indicates the sentences are on different topics and 5 means they are completely equivalent. As our test set, we report the average Pearson's $r$ over each year of the STS tasks from 2012-2016. We use the small (250-example) English dataset from SemEval 2017~\cite{cer2017semeval} as a development set, which we call STS2017 below. 

Section~\ref{sec:supp:paralex} in the appendix contains a description of a method to obtain a paraphrase lexicon from \paranmt that is on par with that provided by PPDB. 
In Section~\ref{sec:general} in the appendix, we also evaluate our sentence embeddings on a range of tasks that have previously been used for evaluating sentence representations~\citep{kiros2015skip}. 

\subsection{Experimental Setup}

For training sentence embeddings on \paranmt, we follow the experimental procedure of \newcite{wieting-16-full}. 
We use \paragramsl embeddings~\cite{wieting2015paraphrase} to initialize the word embedding matrix 
for all models that use word embeddings. We fix the mini-batch size to 100 
and the margin $\delta$ to 0.4. 
We train all models for 5 epochs. For optimization we use Adam~\cite{kingma2014adam} with a learning rate of 0.001. For the LSTM and BLSTM, we fixed the scrambling rate to 0.3.\footnote{Like \citet{wieting-17-full}, we found that scrambling significantly improves results, even though we use much more training data than they used. But while they used a scrambling rate of 0.5, we found that a smaller rate of 0.3 worked better, presumably due to the larger training set.}

\subsection{Dataset Comparison}
\label{sec:data-comp}
We first compare parallel data sources. We evaluate the quality of a data source by using its back-translations paired with its English references as training data for paraphrastic sentence embeddings. We compare the four data sources described in Section~\ref{sec:data}. We use 100K samples from each corpus and trained 3 different models on each: \wordavg, \triavg, and \lstmavg. Table~\ref{table:CorpusSResults} shows that CzEng provides the best training data for all models, so we use it to create \paranmt and in all remaining experiments.

\begin{table}[t]
\setlength{\tabcolsep}{4pt}
\centering
\small
\begin{tabular} { | l || c | c | c |}
\hline
Training Corpus & \wordavg & \triavg & \lstmavg \\
\hline
Common Crawl & 80.9 & 80.2 & 79.1\\
CzEng 1.6 & \bf 83.6 & \bf 81.5 & \bf 82.5\\
Europarl & 78.9 & 78.0 & 80.4\\
News Commentary & 80.2 & 78.2 & 80.5\\
\hline
\end{tabular}
\caption{\label{table:CorpusSResults}
Pearson's $r\times 100$ on STS2017 when training on 100k pairs from each back-translated parallel corpus. CzEng works best for all models.
}
\end{table}

CzEng is diverse in terms of vocabulary and has highly-diverse sentence structures. It has significantly shorter sentences than the other corpora, and has much more training data, so its translations are expected to be better than those in the other corpora. 
\citet{wieting2017backtrans} found that 
sentence length was the most important factor in filtering quality training data, presumably due to how NMT quality deteriorates with longer sentences. We suspect that better translations yield better data for training sentence embeddings. 

\subsection{Data Filtering} \label{filtered-data}

Since the \paranmt dataset is so large, it is computationally demanding to train sentence embeddings on it in its entirety. So, we filter the data to create a training set for sentence embeddings. 

We experiment with three simple methods. We first try using the length-normalized translation score from decoding. 
Second, we use trigram overlap filtering as done by \citet{wieting2017backtrans}.\footnote{Trigram overlap is calculated by counting trigrams in the reference and translation, then dividing the number of shared trigrams by the total number in the reference or translation, whichever has fewer.}
Third, we use the paraphrase score from Section~\ref{sec:data}. 

We filtered the back-translated CzEng data using these three strategies. We ranked all 51M+ paraphrase pairs in the dataset by the filtering measure under consideration and then split the data into tenths (so the first tenth contains the bottom 10\% under the filtering criterion, the second contains those in the bottom 10-20\%, etc.). 

We trained \wordavg, \triavg, and \lstmavg models for a single epoch on 1M examples sampled from each of the ten folds for each filtering criterion. We  averaged the correlation on the STS2017 data across models for each fold. 
Table~\ref{table:STSresultsFold} shows the results of the filtering methods. Filtering based on \paragramphrase similarity produces the best data for training sentence embeddings. 

\begin{table}[h]
\setlength{\tabcolsep}{4pt}
\centering
\small
\begin{tabular} { | l || c |}
\hline
Filtering Method & Model Avg. \\ 
\hline
Trigram Overlap & 83.1 \\
Translation Score & 83.2 \\
Para. Score & \bf 83.3 \\
\hline
\end{tabular}
\caption{\label{table:STSresultsFold}
Pearson's $r\times 100$ on STS2017 for the best training fold across the average of \wordavg, \triavg, and \lstmavg models for each filtering method.
}
\end{table}

We randomly selected 5M examples from the top two scoring folds using \paragramphrase filtering, ensuring that we only selected examples in which both sentences have a maximum length of 30 tokens.\footnote{\citet{wieting2017backtrans} investigated methods to filter back-translated parallel text. They found that sentence length cutoffs were effective for filtering.} 
These resulting 5M examples form the training data for the rest of our experiments. Note that many more than 5M pairs from the dataset are useful, as suggested by our human evaluations in Section~\ref{sec:human}. We have experimented with doubling the training data when training our best sentence similarity model and found the correlation increased by more than half a percentage point on average across all datasets. 

\subsection{Effect of Mega-Batching} \label{sec:megabatch}

Table~\ref{table:STSresultsMB} shows the impact of varying the mega-batch size $M$ when training for 5 epochs on our 5M-example training set. For all models, larger mega-batches improve performance. There is a smaller gain when moving from 20 to 40, but all models show clear gains over $M=1$. 
 
\begin{table}[h]
\setlength{\tabcolsep}{4pt}
\centering
\small
\begin{tabular} { | r || c | c | c |} 
\hline
$M$ & \wordavg & \triavg & \lstmavg \\ 
\hline
1 & 82.3 & 81.5 & 81.5\\
20 & 84.0 & 83.1 & 84.6\\
40 & \bf 84.1 & \bf 83.4 & \bf 85.0\\
\hline
\end{tabular}
\caption{\label{table:STSresultsMB}
Pearson's $r\times 100$ on STS2017 
with different mega-batch sizes $M$. 
}
\end{table}

\begin{table}[h]
\setlength{\tabcolsep}{4pt}
\centering
\small
\begin{tabular}{|ll|} 
\hline
sentence: & sir, i'm just trying to protect. \\
\hline
\multicolumn{2}{|l|}{\bf negative examples:} \\
$M=1$ &i mean, colonel... \\
$M=20$ &i only ask that the baby be safe. \\
$M=40$ &just trying to survive. on instinct. \\
\hline
\hline
sentence: & i'm looking at him, you know? \\
\hline
$M=1$ &they know that i've been looking for her. \\
$M=20$ &i'm keeping him. \\
$M=40$ &i looked at him with wonder. \\
\hline
\hline
sentence: & i'il let it go a couple of rounds. \\
\hline
$M=1$ &sometimes the ball doesn't go down. \\
$M=20$ &i'll take two. \\
$M=40$ &i want you to sit out a couple of rounds, all right? \\
\hline
\end{tabular}
\caption{\label{table:NearestNeighbors}
Negative examples for various mega-batch sizes $M$ with the \blstmavg model.
}
\end{table}

\begin{table*}[th!]
\setlength{\tabcolsep}{6pt}
\small
\centering
\begin{tabular} { |l|l| l | c || c | c | c | c | c |} 
\cline{2-9}
\multicolumn{1}{c|}{} & Training Data & Model & Dim. & 2012 & 2013 & 2014 & 2015 & 2016  \\
\hline
&&\wordavg & 300 & 66.2 & 61.8 & 76.2 & 79.3 & 77.5 \\
&&\triavg & 300 & 67.2 & 60.3 & 76.1 & 79.7 & \bf 78.3 \\
&&\lstmavg & 300 & 67.0 & 62.3 & 76.3 & 78.5 & 76.0  \\
\cline{3-9}
&&\lstmavg & 900 & \bf 68.0 & 60.4 & 76.3 & 78.8 & 75.9 \\
\multicolumn{1}{|l|}{} & \paranmtnonum &\blstmavg & 900 & 67.4 & 60.2 & 76.1 & 79.5 & 76.5 \\
\cline{3-9}
\textbf{Our Work} &&\wordavg+ \triavg (addition) & 300 & 67.3 & \bf 62.8 & \bf 77.5 & 80.1 & 78.2 \\
&&\wordavg+ \triavg+ \lstmavg (addition) & 300 & 67.1 & \bf 62.8 & 76.8 & 79.2 & 77.0 \\
&&\bf \wordavg\concat \triavg (concatenation) & 600 & 67.8 & 62.7 & 77.4 & \bf 80.3 & 78.1 \\
&&\wordavg\concat \triavg\concat \lstmavg (concatenation) & 900 & 67.7 & \bf 62.8 & 76.9 & 79.8 & 76.8 \\
\cline{2-9}
& \multicolumn{1}{l|}{SimpWiki}
&\wordavg\concat \triavg (concatenation) & 600 & 61.8 & 58.4 & 74.4 & 77.0 & 74.0 \\
\hline
\multicolumn{2}{|l|}{}
&$1^{st}$ Place System & - & 64.8 & 62.0 & 74.3 & 79.0 & 77.7 \\
\multicolumn{2}{|l|}{\bf STS Competitions}
&$2^{nd}$ Place System & - & 63.4 & 59.1 & 74.2 & 78.0 & 75.7 \\
\multicolumn{2}{|l|}{}
&$3^{rd}$ Place System & - & 64.1 & 58.3 & 74.3 & 77.8 & 75.7 \\
\hline
\multicolumn{2}{|l|}{}
&InferSent (AllSNLI) \cite{conneau2017supervised} & 4096 & 58.6 & 51.5 & 67.8 & 68.3 & 67.2 \\
\multicolumn{2}{|l|}{}
&InferSent (SNLI) \cite{conneau2017supervised} & 4096 & 57.1 & 50.4 & 66.2 & 65.2 & 63.5 \\
\multicolumn{2}{|l|}{}
&FastSent \cite{hill2016learning} & 100 & - & - & 63 & - & - \\
\multicolumn{2}{|l|}{}
&DictRep \cite{hill2016learning} & 500 & - & - & 67 & - & - \\
\multicolumn{2}{|l|}{\bf Related Work} 
&SkipThought \cite{kiros2015skip} & 4800 & - & - & 29 & - & - \\
\multicolumn{2}{|l|}{}
&CPHRASE \cite{pham-EtAl:2015:ACL-IJCNLP} & - & - & - & 65 & - & - \\
\multicolumn{2}{|l|}{}
&CBOW (from \citealp{hill2016learning}) & 500 & - & - & 64 & - & - \\
\multicolumn{2}{|l|}{}
&BLEU \cite{papineni2002bleu} & - & 39.2 & 29.5 & 42.8 & 49.8 & 47.4 \\
\multicolumn{2}{|l|}{}
&METEOR \cite{denkowski:lavie:meteor-wmt:2014} & - & 53.4 & 47.6 & 63.7 & 68.8 & 61.8 \\
\hline
\end{tabular}
\caption{\label{table:STSresultsSimple}
Pearson's $r\times 100$ on the STS tasks of our models and those from related work. We compare to the top performing systems from each SemEval STS competition. Note that we are reporting the mean correlations over domains for each year rather than weighted means as used in the competitions. Our best performing overall model (\wordavg\concat \triavg) is in bold.
}
\end{table*}

\begin{table}[th!]
\setlength{\tabcolsep}{4pt}
\small
\centering
\begin{tabular} {|lcc|} 
\cline{2-3}
\multicolumn{1}{l|}{} & Dim. & Corr. \\
\hline
\multicolumn{3}{|l|}{\bf Our Work (Unsupervised)} \\
\hline
\wordavg & 300 & 79.2 \\
\triavg & 300 & 79.1 \\
\lstmavg & 300 & 78.4 \\
\hline
\wordavg+ \triavg (addition) & 300 & 79.9 \\
\wordavg+ \triavg+ \lstmavg (addition) & 300 & 79.6 \\
\bf \wordavg\concat \triavg (concatenation) & 600 & 79.9 \\
\wordavg\concat \triavg\concat \lstmavg (concatenation) & 900 & 79.2 \\
\hline
\multicolumn{3}{|l|}{\bf Related Work (Unsupervised)} \\
\hline
InferSent (AllSNLI) \cite{conneau2017supervised} & 4096 & 70.6 \\
C-PHRASE \cite{pham-EtAl:2015:ACL-IJCNLP} && 63.9 \\
GloVe \cite{pennington2014glove} & 300 & 40.6 \\
word2vec \cite{mikolov2013distributed} & 300 & 56.5 \\
sent2vec \cite{pagliardini2017unsupervised} & 700 & 75.5 \\
\hline
\multicolumn{3}{|l|}{\bf Related Work (Supervised)} \\
\hline
Dep. Tree LSTM \cite{tai2015improved} & & 71.2 \\
Const. Tree LSTM \cite{tai2015improved} & & 71.9 \\
CNN \cite{shao2017hcti} && 78.4 \\
\hline
\end{tabular}
\caption{\label{table:STSBenchmark}
Pearson's $r\times 100$ on STS Benchmark test set. 
}
\end{table}

Table~\ref{table:NearestNeighbors} shows negative examples with different mega-batch sizes $M$. 
We use the \blstmavg model and show the negative examples (nearest neighbors from the mega-batch excluding the current training example) for three sentences. Using larger mega-batches improves performance, presumably by producing more compelling negative examples for the learning procedure. 
This is likely more important when training on sentences 
than prior work on learning from text snippets~\cite{wieting2015paraphrase,wieting-16-full,pham-EtAl:2015:ACL-IJCNLP}. 

\subsection{Model Comparison}

\begin{table}
\setlength{\tabcolsep}{4pt}
\small
\centering
\begin{tabular} { | l || c |} 
\hline
Models & Mean Pearson Abs. Diff. \\
\hline
\wordavg~/ \triavg & 2.75 \\
\wordavg~/ \lstmavg & 2.17 \\
\triavg~/ \lstmavg & 2.89 \\
\hline
\end{tabular}
\caption{\label{table:STSdiff}
The means (over all 25 STS competition datasets) of the absolute differences in Pearson's $r$ between each pair of models.
}
\end{table}

\begin{table*}[th!]
    \centering
    \scriptsize
    \begin{tabular}{p{5cm}p{10cm}}
        \normalsize \bf Template & \normalsize \bf Paraphrase \\
        \toprule
          original &  with the help of captain picard, the borg will be prepared for everything. \\
          \texttt{(SBARQ(ADVP)(,)(S)(,)(SQ))} &  now, the borg will be prepared by picard, will it? \\
         \texttt{(S(NP)(ADVP)(VP))} &  the borg here will be prepared for everything. \\
        \midrule
         original &  you seem to be an excellent burglar when the time comes.\\
         \texttt{(S(SBAR)(,)(NP)(VP))}& when the time comes, you'll be a great thief.\\
         \texttt{(S(``)(UCP)('')(NP)(VP))} & ``you seem to be a great burglar, when the time comes.'' you said.\\
         \midrule
         original & overall, i that it's a decent buy, and am happy that i own it. \\
         - & it's a good buy, and i'm happy to own it. \\
         \midrule
         original & oh, that's a handsome women, that is. \\
         - & that's a beautiful woman. \\
         \end{tabular}
        \caption{The top two rows in the table show syntactically controlled paraphrases generated by the SCPN. The bottom two rows are examples from our paraphrase model that are able to canonicalize text and even correct grammar mistakes.}
    \label{table:examples}
\end{table*}

Table~\ref{table:STSresultsSimple} shows the results on the STS tasks from 2012-2016, and Table~\ref{table:STSBenchmark} shows results on the STS Benchmark.\footnote{Baseline results are from \url{http://ixa2.si.ehu.es/stswiki/index.php/STSbenchmark}, except for the unsupervised InferSent result which we computed.} 
Our best models outperform all STS competition systems and all related work of which we are aware on the STS datasets. Note that the large improvement over BLEU and METEOR implies that our embeddings could be useful for evaluating machine translation output.

Overall, our individual models (\wordavg, \triavg, \lstmavg) perform similarly. Using 300 dimensions appears to be sufficient; increasing dimensionality does not necessarily improve correlation. 
When examining particular STS tasks, we found that our individual models showed marked differences on certain tasks. Table~\ref{table:STSdiff} shows the mean absolute difference in Pearson's $r$ over all 25 datasets. 
The \triavg model shows the largest differences from the other two, both of which use word embeddings. 
This suggests that \triavg may be able to complement the other two by providing information about words that are unknown to models that rely on word embeddings. 

We experiment with two ways of combining models. The first is to define additive architectures that form the embedding for a sentence by adding the embeddings computed by two (or more) individual models. 
All parameters are trained jointly just like when we train individual models; that is, we do not first train two simple models and add their embeddings.  
The second way is to define concatenative architectures that form a sentence embedding by concatenating the embeddings computed by individual models, and again to train all parameters jointly. 

In Table~\ref{table:STSresultsSimple} and Table~\ref{table:STSBenchmark}, these combinations show consistent improvement over the individual models as well as the larger \lstmavg and \blstmavg. Concatenating \wordavg and \triavg results in the best performance on average across STS tasks, outperforming the best supervised systems from each year. We will release the pretrained model for these ``\wordavg\concat \triavg'' embeddings upon publication. In addition to providing a strong baseline for future STS tasks, our embeddings offer the advantages of being extremely efficient to compute and being robust to unknown words. 

We show the usefulness of \paranmtnonum by also reporting the results of training the ``\wordavg\concat \triavg'' model on SimpWiki, a dataset of aligned sentences from Simple English and standard English Wikipedia~\citep{coster2011simple}. It has been shown useful for training sentence embeddings in past work~\citep{wieting-17-full}. However, Table~\ref{table:STSresultsSimple} shows that training on \paranmtnonum leads to gains in correlation of 3 to 6 points.

\section{Paraphrase Generation} \label{sec:gen}
Besides creating state-of-the-art paraphrastic sentence embeddings, our dataset is useful for paraphrase generation for purposes of augmenting data and creating adversarial examples. The work is described fully in \cite{iyyer2018controlled}, where their model, the Syntactically Controlled Paraphrase Network (SCPN), is trained to generate a paraphrase of a sentence whose constituent structure follows a provided {\it parse template}. These {\it parse templates} are the top two levels of the linearized parse tree (the level immediately below the root along with the root).

We have also found that training an encoder-decoder model on \paranmt can lead to a model that canonicalizes text. For this experiment, we used a bidirectional LSTM~\cite{hochreiter1997long} encoder and a two-layer LSTM decoder with soft attention over the encoded states!\cite{bahdanau2014neural}. The attention computation consists of a bilinear product with a learned parameter matrix.

Table~\ref{table:examples}, shows two examples from each of these models. Notice how the for the SCPN, the transformation preserves the semantics of the sentence while changing its syntax to fit the templates. The latter two examples show the canonicalization effect where the model is able to correct grammatical errors and standardize the output. This canonicalization would be interesting to explore for automatic grammar correction as it does so without any direct supervision. Future work could also use this canonicalization to improve performance of models by standardizing inputs and removing noise from data.

These were the first studies using \paranmt to generate paraphrases, and we believe that \paranmt and future datasets like it, can be used to generate rich paraphrases that improve the performance and robustness of models on a multitude of NLP tasks and leave this future exploration.

\section{Conclusion}
We described the creation of \paranmt, a dataset of more than 50M English sentential paraphrase pairs. We showed how to use \paranmt to train paraphrastic sentence embeddings that outperform supervised systems on STS tasks, as well as how it can be used for generating paraphrases for purposes of data augmentation, robustness, and even grammar correction.

The key advantage of our approach is that it 
only requires parallel text. 
There are hundreds of millions of parallel sentence pairs, and more are being generated continually. 
Our procedure is immediately applicable to the wide range of languages for which we have parallel text. 

We release \paranmt, our code, and pretrained sentence embeddings, which also exhibit strong performance as general-purpose representations for a multitude of tasks.\footnote{We will release code and embeddings under the permissive MIT and CC BY 4.0 licenses. We will work with the CzEng developers to release \paranmt under the most permissive license possible, but as CzEng is only available for non-commercial research purposes, we may be restricted to release \paranmt under the same license as CzEng.} 
We hope that \paranmt, along with our embeddings, can impart a notion of meaning equivalence to improve NLP systems for a variety of tasks. 
We are actively investigating ways to apply these two new resources to downstream applications, including machine translation, question answering, and paraphrase generation for data augmentation and finding adversarial examples.

\section*{Acknowledgments}

We thank the developers of Theano~\cite{2016arXiv160502688short}, the developers of PyTorch~\cite{paszke2017automatic}, and NVIDIA Corporation for donating GPUs used in this research.

\bibliography{emnlp2017}
\bibliographystyle{acl_natbib_new}

\appendix

\section{Appendix}
\label{sec:supplementary}
\subsection{Paraphrase Lexicon}
\label{sec:supp:paralex}

\begin{table*}[t]
\setlength{\tabcolsep}{4pt}
\small
\centering
\begin{tabular} { | l | l | p{12cm}|} 
\hline
 & PPDB & giggled, smiled, funny, used, grew, bust, ri, did \\
\cline{2-3}
laughed & 
\multirow{2}{*}{\paranmt} & chortled, guffawed, pealed, laughin, laughingstock, cackled, chuckled, snickered, mirthless, chuckling, jeered, laughs, laughing, taunted, burst, cackling, scoffed, humorless, barked,...\\
\hline
\hline
\multirow{4}{*}{respectful} & \multirow{2}{*}{PPDB} & respect, respected, courteous, disrespectful, friendly, respecting, respectable, humble, environmentally-friendly, child-friendly, dignified, respects, compliant, sensitive, abiding,...\\
\cline{2-3}
& \multirow{2}{*}{\paranmt} & reverent, deferential, revered, respectfully, awed, respect, respected, respects, respectable, politely, considerate, treat, civil, reverence, polite, keeping, behave, proper, dignified, decent,... \\
\hline
\end{tabular}
\caption{\label{table:paralex-examples}
Example lexical paraphrases from PPDB ranked using the PPDB 2.0 scoring function and from the paraphrase lexicon we induced from \paranmt ranked using adjusted PMI.
}
\end{table*}

While \paranmt consists of sentence pairs, we demonstrate how a paraphrase lexicon can be extracted from it. One simple approach is to extract and rank word pairs $\langle u, v\rangle$ using the \textbf{cross-sentence pointwise mutual information (PMI)}: 
\[\mathrm{PMI}_{\mathrm{cross}}(u,v) = \log \frac{\#(u,v) \#(\cdot,\cdot)}{\#(u)\#(v)}\]
\noindent where joint counts $\#(u,v)$ are incremented when $u$ appears in a sentence and $v$ appears in its paraphrase. The marginal counts (e.g., $\#(u)$) are computed based on single-sentence counts, as in ordinary PMI. 
This works reasonably well but is not able to differentiate words that frequently occur in paraphrase pairs from words that simply occur frequently together in the same sentence. For example, ``Hong'' and ``Kong'' have high cross-sentence PMI. We can improve the score by subtracting the ordinary PMI that computes joint counts based on single-sentence co-occurrences. We call the result the \textbf{adjusted PMI}:
\[\mathrm{PMI}_{\mathrm{adj}}(u,v) =  \mathrm{PMI}_{\mathrm{cross}}(u,v) - \mathrm{PMI}(u,v)\]
\noindent Before computing these PMIs from \paranmt, we removed sentence pairs with a paraphrase score less than 0.35 and where either sentence is longer than 30 tokens. When computing the ordinary PMI with single-sentence context, we actually compute separate versions of this PMI score for translations and references in each \paranmt pair, then we average them together. We did this because the two sentences in each pair have highly correlated information, so computing PMI on each half of the data would correspond to capturing natural corpus statistics in a standard application of PMI. 

\begin{table}[th!]
\setlength{\tabcolsep}{4pt}
\small
\centering
\begin{tabular} { | l l | c|} 
\hline
Dataset & Score & $\rho\times 100$ \\
\hline
PPDB L & PPDB 2.0 & 37.97 \\
PPDB XL & PPDB 2.0 & 52.32 \\ 
PPDB XXL & PPDB 2.0 & 60.44 \\ 
PPDB XXXL & PPDB 2.0 & 61.47 \\ 
\hline
\paranmt & cross-sentence PMI & 52.12 \\
\paranmt & adjusted PMI & \textbf{61.59} \\
\hline
\end{tabular}
\caption{\label{table:paralex}
Evaluation of scored paraphrase lexicons on SimLex-999, showing Spearman's $\rho\times 100$. 
}
\end{table}
Table~\ref{table:paralex} shows an evaluation of the resulting score functions on the SimLex-999 word similarity dataset~\citep{HillRK14}. As a baseline, we use the lexical portion of PPDB 2.0~\citep{pavlick2015ppdb}, evaluating its ranking score as a similarity score and assigning a similarity of 0 to unseen word pairs.\footnote{If both orderings for a SimLex word pair appear in PPDB, we average their PPDB 2.0 scores. If multiple lexical entries are found with different POS tags, we take the first instance.} Our adjusted PMI computed from \paranmt is on par with the best PPDB lexicon. 

Table~\ref{table:paralex-examples} shows examples from PPDB and our paraphrase lexicon computed from \paranmt. Paraphrases from PPDB are ordered by the PPDB 2.0 scoring function. Paraphrases from our lexicon are ordered using our adjusted PMI scoring function; we only show paraphrases that appeared at least 10 times in \paranmt. 

\subsection{General-Purpose Sentence Embedding Evaluations}
\label{sec:general}

\begin{table*}[th!]
\resizebox{1\linewidth}{!}{
\begin{tabular}{lc|cccc@{\,\,}c@{\,\,}cccc@{}}
\hline
\bf Model & \bf Dim. & \bf MR & \bf CR & \bf SUBJ & \bf MPQA & \bf SST & \bf TREC & \bf MRPC & \bf SICK-R & \bf SICK-E \\
\hline
\multicolumn{11}{|l|}{\bf Unsupervised (Unordered Sentences)} \\
\hline
\multicolumn{2}{l|}{Unigram-TFIDF \cite{hill2016learning}} & 73.7 & 79.2 & 90.3 & 82.4 & - & 85.0 & 73.6/81.7 & - & - \\
{SDAE \cite{hill2016learning}} & 2400 & 74.6 & 78.0 & 90.8 &  86.9 & - & 78.4 & 73.7/80.7 & - & - \\
\hline
\multicolumn{11}{|l|}{\bf Unsupervised (Ordered Sentences)} \\
\hline
{FastSent \cite{hill2016learning}} & 100 & 70.8 & 78.4 & 88.7 & 80.6 & - & 76.8 & 72.2/80.3 & - & - \\
\multicolumn{2}{l|}{FastSent+AE \cite{hill2016learning}} & 71.8 & 76.7 & 88.8 & 81.5 & - & 80.4 & 71.2/79.1 & - & - \\
{SkipThought \cite{kiros2015skip}} & 4800 & 76.5 & 80.1 & 93.6 & 87.1 & 82.0 & 92.2 & 73.0/82.0 & 85.8 & 82.3 \\
\hline
\multicolumn{11}{|l|}{\bf Unsupervised (Structured Resources)} \\
\hline
{DictRep \cite{hill2016learning}} & 500 & 76.7 & 78.7 & 90.7 & 87.2 & - & 81.0 & 68.4/76.8 & - & - \\
{NMT En-to-Fr \cite{hill2016learning}} & 2400 &  64.7 &  70.1 & 84.9 &  81.5 &  - & 82.8 & - \\
{BYTE mLSTM \cite{radford2017learning}} & 4096 &\bf 86.9 &\bf 91.4 & 94.6 & 88.5 & - & - & 75.0/82.8 & 79.2 & - \\
\hline
\multicolumn{11}{|l|}{\it Individual Models (Our Work)} \\
\hline
\wordavg & 300 & 75.8 & 80.5 & 89.2 & 87.1 & 80.0 & 80.1 & 68.6/80.9 & 83.6 & 80.6 \\
\triavg & 300 & 68.8 & 75.5 & 83.6 & 82.3 & 73.6 & 73.0 & 71.4/82.0 & 79.3 & 78.0 \\
\lstmavg & 300 & 73.8 & 78.4 & 88.5 & 86.5 & 80.6 & 76.8 & 73.6/82.3 & 83.9 & 81.9 \\
\hline
\lstmavg & 900 & 75.8 & 81.7 & 90.5 & 87.4 & 81.6 & 84.4 & 74.7/83.0 & 86.0 & 83.0 \\
\blstmavg & 900 & 75.6 & 82.4 & 90.6 & 87.7 & 81.3 & 87.4 & 75.0/82.9 & 85.8 & 84.4 \\
\hline
\multicolumn{11}{|l|}{\it Mixed Models (Our Work)} \\
\hline
\wordavg + \triavg (addition) & 300 & 74.8 & 78.8 & 88.5 & 87.4 & 78.7 & 79.0 & 71.4/81.4 & 83.2 & 80.6 \\
\wordavg + \triavg + \lstmavg (addition) & 300 & 75.0 & 80.7 & 88.6 & 86.6 & 77.9 & 78.6 & 72.7/80.8 & 83.6 & 81.8 \\
\wordavg\concat \triavg (concatenation) & 600 & 75.8 & 80.5 & 89.9 & 87.8 & 79.7 & 82.4 & 70.7/81.7 & 84.6 & 82.0 \\
\wordavg\concat \triavg\concat \lstmavg (concatenation) & 900 & 77.6 & 81.4 & 91.4 & 88.2 & 82.0 & 85.4 & 74.0/81.5 & 85.4 & 83.8 \\
\hline
\blstmavg (Avg., concatenation) & 4096 & 77.5 & 82.6 & 91.0 & 89.3 & 82.8 & 86.8 & 75.8/82.6 & 85.9 & 83.8 \\
\blstmavg (Max, concatenation) & 4096 & 76.6 & 83.4 & 90.9 & 88.5 & 82.0 & 87.2 & 76.6/83.5 & 85.3 & 82.5 \\
\hline
\multicolumn{11}{|l|}{\bf Supervised (Transfer)} \\
\hline
InferSent (SST) \cite{conneau2017supervised} & 4096 & - & 83.7 & 90.2 & 89.5 & - & 86.0 & 72.7/80.9 & 86.3 & 83.1 \\
InferSent (SNLI) \cite{conneau2017supervised} & 4096 & 79.9 & 84.6 & 92.1 & 89.8 & 83.3 & 88.7 & 75.1/82.3 &\bf 88.5 &\bf 86.3 \\
InferSent (AllNLI) \cite{conneau2017supervised} & 4096 & 81.1 & 86.3 & 92.4 & 90.2 & 84.6 & 88.2 & 76.2/83.1 & 88.4 &\bf 86.3 \\
\hline
\multicolumn{11}{|l|}{\bf Supervised (Direct)} \\
\hline
\multicolumn{2}{l|}{Naive Bayes - SVM}  & 79.4 & 81.8 & 93.2 & 86.3 &  83.1 & - & - & - & - \\
\multicolumn{2}{l|}{AdaSent \cite{zhao2015self}} &  83.1 & 86.3 &\bf 95.5 &\bf 93.3 & - & 92.4 & - & - & - \\
\multicolumn{2}{l|}{BLSTM-2DCNN~\citep{zhou-EtAl:2016:COLING2}} & 82.3 & - & 94.0 & - &\bf 89.5 &\bf 96.1 & - & - & - \\
\multicolumn{2}{l|}{TF-KLD \cite{ji2013discriminative}} &  - & - & - & - & - & - &\bf 80.4/85.9 & - & - \\
\multicolumn{2}{l|}{Illinois-LH \cite{lai2014illinois}} &  - & - & - & - & - & - & - & - & 84.5 \\
\multicolumn{2}{l|}{Dependency Tree-LSTM \cite{tai2015improved}} &  - & - & - & - & - & - & - & 86.8 & - \\
\hline
\end{tabular}
}
\caption{General-purpose sentence embedding tasks, divided into categories based on resource requirements. 
\label{table:tasks_results}
}
\end{table*}

We evaluate our sentence embeddings on a range of tasks that have previously been used for evaluating sentence representations~\citep{kiros2015skip}. These include sentiment analysis (MR, \citealp{pang2005seeing}; CR, \citealp{hu2004mining}; SST, \citealp{socher2013recursive}), subjectivity classification 
(SUBJ; \citealp{pang2004sentimental}), opinion polarity (MPQA; \citealp{wiebe2005annotating}), question classification (TREC; \citealp{li2002learning}), paraphrase detection (MRPC; \citealp{dolan2004unsupervised}), semantic relatedness (SICK-R; \citealp{marelli2014semeval}), and textual entailment (SICK-E). We use the SentEval package from \citet{conneau2017supervised} to train models
on our fixed sentence embeddings for each task.\footnote{Available at \url{https://github.com/facebookresearch/SentEval}.}

Table~\ref{table:tasks_results} shows results on the general sentence embedding tasks. 
Each of our individual models produces 300-dimensional sentence embeddings, which is far fewer than the several thousands (often 2400-4800) of dimensions used in most prior work.
While using higher dimensionality does not improve correlation on the STS tasks, it does help on the general sentence embedding tasks. Using higher dimensionality leads to more trainable parameters in the subsequent classifiers, increasing their ability to linearly separate the data. 

To enlarge the dimensionality, we concatenate the forward and backward states prior to averaging. 
This is similar to \citet{conneau2017supervised}, though they used max pooling. 
We experimented with both averaging (``\blstmavg (Avg., concatenation)'')~and max pooling (``\blstmavg (Max, concatenation)'')~using recurrent networks with 2048-dimensional hidden states, so concatenating them yields a 
4096-dimension embedding. 
These high-dimensional models 
outperform SkipThought~\citep{kiros2015skip} on all tasks except SUBJ and TREC. Nonetheless, the InferSent~\cite{conneau2017supervised} embeddings trained on AllNLI still outperform our embeddings on nearly all of these general-purpose tasks. 

We also note that on five tasks (SUBJ, MPQA, SST, TREC, and MRPC), all sentence embedding methods are outperformed by supervised baselines. These baselines use the same amount of supervision as the general sentence embedding methods; the latter actually use far more data overall than the supervised baselines. This suggests that the pretrained sentence representations are not capturing the features learned by the models engineered for those tasks. 

We take a closer look of how our embeddings compare to InferSent~\citep{conneau2017supervised}. InferSent is a supervised model trained on a large textual entailment dataset (the SNLI and MultiNLI corpora~\cite{bowman2015large, williams2017broad}, which consist of nearly 1 million human-labeled examples). 

While InferSent has strong performance across all downstream tasks, our model obtains better results on semantic similarity tasks. 
It consistently reach correlations approximately 10 points higher than those of InferSent. 

Regarding the general-purpose tasks, we note that some result trends appear to be influenced by the domain of the data. InferSent is trained on a dataset of mostly captions, especially the model trained on just SNLI. Therefore, the datasets for the SICK relatedness and entailment evaluations are similar in domain to the training data of InferSent. Further, the training task of natural language inference is aligned to the SICK entailment task. Our results on MRPC and entailment are significantly better than SkipThought, and on a paraphrase task that does not consist of caption data (MRPC), our embeddings are competitive with InferSent. To quantify these domain effects, we performed additional experiments that are described in Section~\ref{sec:supp:infersent}. 

There are many ways to train sentence embeddings, each with its own strengths. 
InferSent, our models, and the BYTE mLSTM of \citet{radford2017learning} each excel in particular classes of downstream tasks. 
Ours are specialized for semantic similarity. BYTE mLSTM is trained on review data and therefore is best at the MR and CR tasks. Since the InferSent models are trained using entailment supervision and on caption data, they excel on the SICK tasks. Future work will be needed to combine multiple supervision signals to generate embeddings that perform well across all tasks.

\subsubsection{Effect of Training Domain on InferSent}
\label{sec:supp:infersent}
We performed additional experiments to investigate the impact of training domain on downstream tasks. We first compare the performance of our ``\wordavg\concat \triavg (concatenation)'' model to the InferSent SNLI and AllNLI models on all STS tasks from 2012-2016. We then compare the overall mean with that of the three caption STS datasets within the collection. The results are shown in Table~\ref{table:STSresultsCap}. 
The InferSent models are much closer to our \wordavg\concat \triavg model on the caption datasets than overall, and InferSent trained on SNLI shows the largest difference between its overall performance and its performance on caption data.

\begin{table}[th!]
\setlength{\tabcolsep}{4pt}
\centering
\small
\begin{tabular} { | l || c | c |} 
\hline
Data & AllNLI & SNLI \\
\hline
Overall mean diff. & 10.5 & 12.5 \\
\hline
MSRvid (2012) diff. & 5.2 & 4.6 \\
Images (2014) diff. & 6.4 & 4.8 \\
Images (2015) diff. & 3.6 & 3.0 \\ 
\hline
\end{tabular}
\caption{\label{table:STSresultsCap}
Difference in correlation (Pearson's $r\times 100$) between ``\wordavg\concat \triavg'' and InferSent models trained on two different datasets: AllNLI and SNLI. The first row is the mean difference across all 25 datasets, then the following rows show differences on three individual datasets that are comprised of captions. The InferSent models are much closer to our model on the caption datasets than overall.}
\end{table}

\begin{table}
\setlength{\tabcolsep}{4pt}
\centering
\small
\begin{tabular} { | l || c | c | c |} 
\hline
Model & All & Cap. & No Cap. \\
\hline
\multicolumn{4}{|l|}{\bf Unsupervised} \\
\hline
InferSent (AllNLI) & 70.6 & 83.0 & 56.6 \\
InferSent (SNLI) & 67.3 & 83.4 & 51.7 \\
\wordavg\concat \triavg & 79.9 & 87.1 & 71.7 \\
\hline
\multicolumn{4}{|l|}{\bf Supervised} \\
\hline
InferSent (AllNLI) & 75.9 & 85.4 & 64.8 \\
InferSent (SNLI) & 75.9 & 86.4 & 63.1 \\
\hline
\end{tabular}
\caption{\label{table:STSresultsBench}
STS benchmark results (Pearson's $r\times 100$) comparing our \wordavg, \triavg model to InferSent trained on AllNLI and SNLI. 
Unsupervised results were obtained by simply using cosine similarity of the pretrained embeddings on the test set with no training or tuning. 
Supervised results were obtained by training and tuning using the training and development data of the STS Benchmark. 
We report results using all of the data (All), only the caption portion of the data (Cap.), and all of the data except for the captions (No Cap.).
}
\end{table}

We also compare the performance of these models on the STS Benchmark under several conditions (Table~\ref{table:STSresultsBench}). We first compare unsupervised results on the entire test set, the subset consisting of captions (3,250 of the 8,628 examples in the test set), and the remainder. We include analogous results in the supervised setting, where we filter the respective training and development sets in addition to the test sets. Compared to our model, InferSent shows a much larger gap between captions and non-captions, providing evidence of a bias. Note that this bias is smaller for the model trained on AllNLI, as its training data includes other domains.

\end{document}